\documentclass[11pt]{llncs}
\usepackage{makeidx}  
\begin{document}
\mainmatter              
\title{Sentiment Analysis of Citations Using Word2vec}
\titlerunning{Sentiment Analysis of Citations Using Word2vec}  
%
\author{Haixia Liu}
\institute{
School Of Computer Science, University of Nottingham Malaysia Campus, Jalan Broga, 43500 Semenyih, Selangor Darul Ehsan.\\
\email{khyx3lhi@nottingham.edu.my}\\ 
}

\maketitle              

\begin{abstract}        
Citation sentiment analysis is an important task in scientific paper analysis. Existing machine learning techniques for citation sentiment analysis are focusing on labor-intensive feature engineering, which requires large annotated corpus. As an automatic feature extraction tool, word2vec has been successfully applied to sentiment analysis of short texts. In this work, I conducted empirical research with the question: how well does word2vec work on the sentiment analysis of citations? The  proposed method constructed sentence vectors (sent2vec) by averaging the word embeddings, which were learned from Anthology Collections (ACL-Embeddings). I also investigated polarity-specific word embeddings (PS-Embeddings) for classifying positive and negative citations. The sentence vectors formed a feature space, to which the examined citation sentence was mapped to. Those features were input into classifiers (support vector machines) for supervised classification. Using 10-cross-validation scheme, evaluation was conducted on a set of annotated citations. The results showed that word embeddings are effective on classifying positive and negative citations. However, hand-crafted features performed better for the overall classification. 
\end{abstract}
\section{Introduction}

The evolution of scientific ideas happens when old ideas are replaced by new ones. Researchers usually conduct scientific experiments based on the previous publications. They either take use of others work as a solution to solve their specific problem, or they improve the results documented in the previous publications by introducing new solutions. I refer to the former as positive citation and the later negative citation. Citation sentence examples \footnote{\footnotesize{\tt Randomly selected from : http://cl.awaisathar.com\\/citation-sentiment-corpus/}} with different sentiment polarity are shown in Table \ref{tab:sentimentexamples}. 

\begin{table}
\centering
\small
\begin{tabular}{cccc}
\begin{tabular}{|p{1cm}|p{1cm}|p{3cm}|p{7cm}|}
\hline
{\bf Citing} & {\bf Cited} & {\bf Polarity} & {\bf Examples}\\\hline
{A1} & {A0} & {Positive} & {One of the most effective taggers based on a pure HMM is that developed at Xerox (Cutting \textit{et al.} , 1992).} \\\hline
{A2} & {A0} & {Negative} & {Brill's results demonstrate that this approach can outperform the Hidden Markov Model approaches that are frequently used for part-of-speech tagging (Jelinek, 1985; Church, 1988; DeRose, 1988; Cutting \textit{et al.} , 1992; Weischedel \textit{et al.}, 1993), as well as showing promise for other applications.} \\\hline
\end{tabular} & 
\begin{tabular}{|l|l|l|l|}
\hline
\end{tabular}
\end{tabular}
\caption{Examples of positive and negative citations.}\label{tab:sentimentexamples}
\end{table}

Sentiment analysis of citations plays an important role in plotting scientific idea flow. I can see from Table \ref{tab:sentimentexamples}, one of the ideas introduced in paper A0 is \textit{Hidden Markov Model (HMM) based part-of-speech (POS) tagging}, which has been referenced positively in paper A1. In paper A2, however, a better approach was brought up making the idea (\textit{HMM based POS}) in paper A0 negative. This citation sentiment analysis could lead to future-works in such a way that new approaches (mentioned in paper A2) are recommended to other papers which cited A0 positively \footnote{\footnotesize{Restriction: the citations share the similar topics.\\ In this case: \textit{HMM based POS tagging}}}. Analyzing citation sentences during literature review is time consuming. Recently, researchers developed algorithms to automatically analyze citation sentiment. For example, 
\cite{abu2013purpose} extracted several features for citation purpose and polarity classification, such as reference count, contrary expression and dependency relations. Jochim \textit{et al.} tried to improve the result by using unigram and bigram features \cite{jochim2014improving}. 
\cite{athar2011sentiment} used word level features, contextual polarity features, and sentence structure based features to detect sentiment citations. 
Although they generated good results using the combination of features, it required a lot of engineering work and big amount of annotated data to obtain the features. Further more, capturing accurate features relies on other NLP techniques, such as part-of-speech tagging (POS) and sentence parsing. Therefore, it is necessary to explore other techniques that are free from hand-crafted features. With the development of neural networks and deep learning, it is possible to learn the representations of concepts from unlabeled text corpus automatically. These representations can be treated as concept features for classification. An important advance in this area is the development of the word2vec technique \cite{mikolov2013efficient}, which has proved to be an effective approach in Twitter sentiment classification \cite{tang2014learning}. 

In this work, the word2vec technique on sentiment analysis of citations was explored. Word embeddings trained from different corpora were compared.

\section{Related Work}
Mikolov \textit{et al.} introduced word2vec technique \cite{mikolov2013efficient} that can obtain word vectors by training text corpus. The idea of word2vec (word embeddings) originated from the concept of distributed representation of words \cite{Hinton86}. The common method to derive the vectors is using neural probabilistic language model \cite{bengio2003neural}. Word embeddings proved to be effective representations in the tasks of sentiment analysis \cite{tang2014learning,xue2014study,zhang2015chinese} and text classification \cite{lilleberg2015support}. Sadeghian and Sharafat \cite{sadeghianbag} extended word embeddings to sentence embeddings by averaging the word vectors in a sentiment review statement. Their results showed that word embeddings outperformed the bag-of-words model in sentiment classification. In this work, I are aiming at evaluating word embeddings for sentiment analysis of citations. The  research questions are:
\begin{enumerate} 
\item How well does word2vec work on classifying positive and negative citations?
\item Can sentiment-specific word embeddings improve the classification result?
\item How well does word2vec work on classifying implicit citations?
\item In general, how well does word2vec work on classifying positive, negative and objective citations in comparison with hand-crafted features?
\end{enumerate}
\section{Methodology}
\label{sec:methodology}

\subsection{Pre-processing}\label{sec:Pre-processing}

The SentenceModel provided by LingPipe was used to segment raw text into its constituent sentences \footnote{\footnotesize{\tt http://alias-i.com/lingpipe/docs/api/\\com/aliasi/sentences/SentenceModel.html}}. 
The data I used to train the vectors has noise. For example, there are incomplete sentences mistakenly detected (e.g. Publication Year.). To address this issue, I eliminated sentences with less than three words.

\subsection{Overall Sent2vec Training}\label{sec:word2vectraining}
In the  work, I constructed sentence embeddings based on word embeddings. I simply averaged the vectors of the words in one sentence to obtain sentence embeddings (sent2vec). The main process in this step is to learn the word embedding matrix $W_{w}$:\\

{$V_{sent2vec}(w) = $ $\frac{1}{n}$ $\sum$ $W^{x_{i}}_{w}$} (1)\\

where $W_{w}$ ($w = <w_{1},x_{2},...w_{n}>$) is the word embedding for word $x_{i}$, which could be learned by the classical word2vec algorithm \cite{mikolov2013efficient}. The parameters that I used to train the word embeddings are the same as in the work of Sadeghian and Sharafat 

\subsection{Polarity-Specific Word Representation Training}\label{sec:specword2vectraining}

To improve sentiment citation classification results, I trained polarity specific word embeddings (PS-Embeddings), which were inspired by the Sentiment-Specific Word Embedding \cite{tang2014learning}. After obtaining the PS-Embeddings, I used the same scheme to average the vectors in one sentence according to the  sent2vec model.

\section{Experiment}
\label{sec:Experiment}

\subsection{Training Dataset}\label{sec:trainingdataset}

The ACL-Embeddings (300 and 100 dimensions) from ACL collection were trained . ACL Anthology Reference Corpus \footnote{\footnotesize{\tt http://acl$-$arc.comp.nus.edu.sg/}} contains the canonical 10,921 computational linguistics papers, from which I have generated 622,144 sentences after filtering out sentences with lower quality.  

 

For training polarity specific word embeddings (PS-Embeddings, 100 dimensions), I selected 17,538 sentences (8,769 positive and 8,769 negative) from ACL collection, by comparing sentences with the polar phrases \footnote{\footnotesize{\tt http://cl.awaisathar.com\\/citation-sentiment-corpus/}}.

The pre-trained Brown-Embeddings (100 dimensions) learned from Brown corpus was also used \footnote{\footnotesize{\tt h$ttps://en.wikipedia.org/wiki/\\Brown_Corpus$}} as a comparison.

\subsection{Test Dataset}\label{sec:testdataset}

To evaluate the sent2vec performance on citation sentiment detection, I conducted experiments on three datasets. The first one (dataset-basic) was originally taken from ACL Anthology \cite{bird2008acl}. Athar and Awais \cite{athar2011sentiment} manually annotated 8,736 citations from 310 publications in the ACL Anthology. I used all of the labeled sentences (830 positive, 280 negative and 7,626 objective) for testing. 
\footnote{\footnotesize{In \cite{athar2011sentiment}'s work, they used 244 negative, 743 positive and 6277 objective citations for testing.}}

The second dataset (dataset-implicit) was used for evaluating implicit citation classification, containing 200,222 excluded (x), 282 positive (p), 419 negative (n) and 2,880 objective (o) annotated sentences. Every sentence which does not contain any direct or indirect mention of the citation is labeled as being excluded (x) \footnote{\footnotesize{\tt$http://www.cl.cam.ac.uk/$\textasciitilde$aa496\\/citation$-$context$-$corpus/$}}. 

The third dataset (dataset-pn) is a subset of dataset-basic, containing 828 positive and 280 negative citations. Dataset-pn was used for the purposes of (1) evaluating binary classification (positive versus negative) performance using sent2vec; (2) Comparing the sentiment classification ability of PS-Embeddings with other embeddings.  

\subsection{Evaluation Strategy}\label{sec:results}

One-Vs-The-Rest strategy was adopted \footnote{\footnotesize{\tt $http://scikit-learn.org/stable/\\modules/multiclass.html$}} for the task of multi-class classification and I reported F-score, micro-F, macro-F and weighted-F scores \footnote{\footnotesize{\tt $http://scikit-learn.org/stable/modules/\\generated/sklearn.metrics.f1_score.html$}} using 10-fold cross-validation. The F1 score is a weighted average of the precision and recall. In the multi-class case, this is the weighted average of the F1 score of each class. There are several types of averaging performed on the data: Micro-F calculates metrics globally by counting the total true positives, false negatives and false positives. Macro-F calculates metrics for each label, and find their unweighted mean. Macro-F does not take label imbalance into account. Weighted-F calculates metrics for each label, and find their average, weighted by support (the number of true instances for each label). Weighted-F alters macro-F to account for label imbalance. 

\subsection{Results}\label{sec:results}
The performances of citation sentiment classification on dataset-basic and dataset-implicit were shown in Table \ref{tab:result1} and Table \ref{tab:result2} respectively. The result of classifying positive and negative citations was shown in Table \ref{tab:result3}. To compare with the outcomes in the work of \cite{athar2011sentiment} \footnote{\footnotesize{The test dataset is slightly larger than \cite{athar2011sentiment}'s test dataset.}}, I selected two records from their results: the best one (based on features n-gram + dependencies + negation) and the baseline (based on 1-3 grams). From Table \ref{tab:result1} I can see that the features extracted by 
\cite{athar2011sentiment} performed far better than word embeddings, in terms of macro-F (their best macro-F is 0.90, the one in this work is 0.33). However, the higher micro-F score (The highest micro-F in this work is 0.88, theirs is 0.78) and the weighted-F scores indicated that this method may achieve better performances if the evaluations are conducted on a balanced dataset. Among the embeddings, ACL-Embeddings performed better than Brown corpus in terms of macro-F and weighted-F measurements \footnote{\footnotesize{I did not perform significant test for the comparison.}}. To compare the dimensionality of word embeddings, ACL300 gave a higher micro-F score than ACL100, but there is no difference between 300 and 100 dimensional ACL-embeddings when look at the macro-F and weighted-F scores.
\begin{table}
\centering
\small
\begin{tabular}{|p{3cm}|p{3cm}|p{3cm}|p{3cm}|}
\hline
{\bf Methods } & {\bf Micro-F} & {\bf Macro-F} & {\bf Weigh-F} \\\hline
{ACL300} & {0.88} & {0.33} & {0.82}\\
{ACL100} & {0.87} & {0.33} & {0.82}\\
{Brown100} & {0.87} & {0.31} & {0.81}\\\hline
{n-grams} & {0.60} & {0.87} & {-}\\
{"+dep+neg} & {0.76} & {0.90} & {-}\\\hline
\end{tabular}
\caption{Performance of citation sentiment classification.}\label{tab:result1}
\end{table}

Table \ref{tab:result2} showed the sent2vec performance on classifying implicit citations with four categories: objective, negative, positive and excluded. 
The method in this experiment had a poor performance on detecting positive citations, but it was comparable with both the baseline and sentence structure method \cite{Athar:2012:DIC:2391171.2391176} for the category of objective citations. With respect to classifying negative citations, this method was not as good as sentence structure features
but it outperformed the baseline. The results of classifying category X from the rest showed that the performances of this method and the sentence structure method are fairly equal.

\begin{table}
\centering
\small
\begin{tabular}{|p{3cm}|p{3cm}|p{3cm}|p{3cm}|}
\hline
{\bf Sentiment } & {\bf Baseline} & {\bf Athar} & {\bf ACL300} \\\hline
{O (F-score)} & {0.86} & {0.89} & {0.84} \\
{N (F-score)} & {0.14} & {0.62} & {0.44} \\
{P (F-score)} & {0.40} & {0.55} & {0.27} \\\hline
{Macro-F}& {0.47} & {0.69} & {0.44} \\
{Weighted-F}& {-} & {-} & {0.77} \\
\hline 
{X vs O,N,P (F-score)} & {0.990} & {0.996} & {0.997} \\\hline
\end{tabular}
\caption{Performance of implicit citation sentiment classification.}\label{tab:result2}

\end{table}

Table \ref{tab:result3} showed the results of classifying positive and negative citations using different word embeddings. The macro-F score 0.85 and the weighted-F score 0.86 proved that word2vec is effective on classifying positive and negative citations.  
However, unlike the outcomes in the paper of
\cite{tang2014learning}, where they concluded that sentiment specific word embeddings performed best, integrating polarity information did not improve the result in this experiment. 

\begin{table} 
\centering
\small
\begin{tabular}{|p{3cm}|p{3cm}|p{3cm}|}
\hline
{\bf Trained Corpus } & {\bf Macro-F }& {\bf Weigh-F }\\\hline
{Brown100} & {0.84} & {0.85}\\
{ACL300} & {0.85} & {0.86}\\
{ACL100} & {0.85} & {0.85}\\
{PS-ACL300} & {0.84}& {0.85}\\\hline
\end{tabular}
\caption{Performance of classifying positive and negative citations. 
}\label{tab:result3}
\end{table}

\section{Discussion and Conclusion}
\label{sec:discon}
In this paper, I reported the citation sentiment classification results based on word embeddings. The binary classification results in Table \ref{tab:result3} showed that word2vec is a promising tool for distinguishing positive and negative citations. From Table \ref{tab:result3} I can see that there are no big differences among the scores generated by ACL100 and Brown100, despite they have different vocabulary sizes (ACL100 has 14,325 words, Brown100 has 56,057 words). The polarity specific word embeddings did not show its strength in the task of binary classification. For the task of classifying implicit citations (Table \ref{tab:result2}), in general, sent2vec (macro-F 0.44) was comparable with the baseline (macro-F 0.47) and it was effective for detecting objective sentences (F-score 0.84) as well as separating X sentences from the rest (F-score 0.997), but it did not work well on distinguishing positive citations from the rest. For the overall classification (Table \ref{tab:result1}), however, this method was not as good as hand-crafted features, such as n-grams and sentence structure features. I may conclude from this experiment that word2vec technique has the potential to capture sentiment information in the citations, but hand-crafted features have better performance.
  
\end{document}